\documentclass[letterpaper, 10 pt, conference]{ieeeconf}  

\IEEEoverridecommandlockouts                              

\usepackage{graphicx}
\usepackage{subcaption}

\usepackage{array}
\newcolumntype{P}[1]{>{\centering\arraybackslash}p{#1}}
\newcolumntype{L}[1]{>{\raggedright\let\newline\\\arraybackslash\hspace{0pt}}m{#1}}
\usepackage{multirow}
\usepackage{csquotes}

\title{\LARGE \bf
Autonomous Vehicle Visual Signals for Pedestrians: \\Experiments and Design Recommendations*
}

\author{Henry Chen, Robin Cohen, Kerstin Dautenhahn, Edith Law, Krzysztof Czarnecki$^{1}$
\thanks{*This work was not supported by any organization}
\thanks{$^{1}$Authors are with the Cheriton School of Computer Science or the Department of Electrical and Computer Engineering,
        University of Waterloo, ON N2L 3G1, Canada
        {\tt\small \{henry.chen.1, kczarnec, rcohen, kerstin.dautenhahn, edith.law\}@uwaterloo.ca}}%
}

\begin{document}

\maketitle
\thispagestyle{empty}
\pagestyle{empty}

\begin{abstract}

Autonomous Vehicles (AV) will transform transportation, but also the interaction between vehicles and pedestrians. In the absence of a driver, it is not clear how an AV can communicate its intention to pedestrians. One option is to use visual signals. To advance their design, we conduct four human-participant experiments and evaluate six representative AV visual signals for visibility, intuitiveness, persuasiveness, and usability at pedestrian crossings. Based on the results, we distill twelve practical design recommendations for AV visual signals, with focus on signal pattern design and placement. Moreover, the paper advances the methodology for experimental evaluation of visual signals, including lab, closed-course, and public road tests using an autonomous vehicle. In addition, the paper also reports insights on pedestrian crosswalk behaviours and the impacts of pedestrian trust towards AVs on the behaviors. We hope that this work will constitute valuable input to the ongoing development of international standards for AV lamps, and thus help mature automated driving in general.

\end{abstract}

\section{INTRODUCTION}

The world is entering the autonomous driving era. Ironically, for a technology that promises to enhance traffic safety, there has been little research on how autonomous vehicles can maintain the communication with pedestrians. Visual signals is a promising solution for intent communication~\cite{webSAEJ3134201905}. However, a crucial question remains:~\enquote{which form of visual signals is best for communicating an AV's intention?} 

\begin{figure}[thpb]
  \centering
  \includegraphics[width=0.75\linewidth]{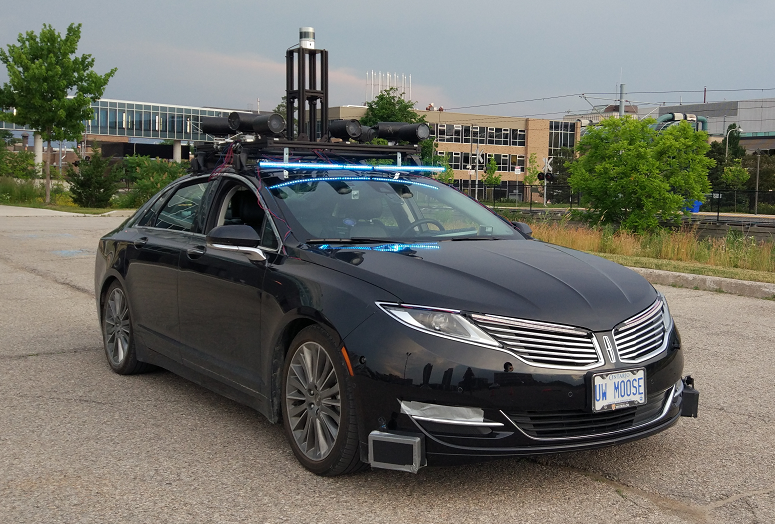}
  \caption{The Autonomoose research vehicle with the visual signal communication system prototype (S-T signal is ON)}
  \label{fig:moose}
\end{figure}

We address this question by exploring how an AV can communicate its intention to yield at the crosswalk. Specifically, we examine the following sub-questions in four experiments (E1-E4). In E1, we explore the visibility of visual signals in human peripheral vision (RQ1) and whether the signal mounting location matters (RQ2). E2 investigates pedestrians' intuitive reaction to signals (RQ3), the effect of training on their reaction (RQ4), and their reaction time (RQ5).  In E3, we deployed a closed-course-experiment featuring a fully automated vehicle, to study the effect of visual signals of an approaching vs. a stationary vehicle (RQ6 and RQ7), the impact of repeat interactions (RQ8), and the impact of trust on crossing behaviours (RQ9). Finally, E4 is a Wizard-of-Oz field study, in which we explore pedestrians' recognition of an autonomous vehicle (RQ10), their recognition and impression of visual signals (RQ11), and the impact of visual signals on pedestrians' crosswalk behaviour (RQ12). Ultimately, this paper makes the following contributions to the fields of Pedestrians Behaviour and Vehicle-Pedestrian Communication: (1) twelve design recommendations for effective AV intention communications, (2) four novel experiment methodologies for evaluating visual signals, and (3) new insights about pedestrian crosswalk behaviours.

\section{Background}
\subsection{Pedestrian Crosswalk Behaviours}
The decision-making process of pedestrians at crosswalks is complex and multifaceted. Rasouli \textit{et al.} summarize eight main factors influencing pedestrian decisions \cite{rasouli2018survey}: (1) physical context, such as traffic signal and lighting conditions; (2) dynamic factors, e.g., vehicle speed, gap acceptance, and communication; (3) traffic characteristics, e.g., traffic volume and law enforcement; (4) social factors, e.g., social norms and group size; (5) demographics, e.g., gender and age; (6) abilities, e.g., speed and distance judgement; (7) personal state, e.g., group size and attention; and finally, (8) personal characteristics, e.g., past experience and cultural background. The survey further argues that there is a lack of studies examining these factors in the context of autonomous driving. Our work contributes new understandings about communication (under dynamic factors), as it explores how to replace driver communication with visual signals. 

\subsection{AV Intention Communication}
Currently, there exists two opposing views about the necessity of AV intention communication. First, some researchers believe that vehicle-pedestrian communication is not necessary because explicit driver communication (e.g., eye contact) is rarely used even in high density urban traffic settings \cite{Nathanael2019}, and that a vehicle's motion and movement is enough to express its intention to pedestrians \cite{Camara2018}. However, other researchers believe it is necessary for AV to explicitly communicate intention. For instance, Gueguen \textit{et al.} argue that eye contact is a powerful form of communication that promotes traffic cooperation and rule compliance \cite{Gueguen2015} and Lundgren \textit{et al.} argue failing to communicate intention will lead to a decrease of pedestrian confidence and trust of AVs \cite{Lundgren2017}. Despite the controversy, the automotive industry now agrees that intention communication is necessary. Specifically, SAE International and the International Organization for Standardization (ISO) recently released the standards of \enquote{Automated Driving System Marker Lamp (J3134) \cite{webSAEJ3134201905}} and \enquote{Ergonomic Aspects of External Visual Communication From Automated Vehicles to Other Road Users (ISO/TR 23049) \cite{webISO230492018},} and the ISO is also developing the two additional standards:~\enquote{Ergonomic Design Guidance for External Visual Communication From Automated Vehicles to Other Road Users (ISO/NP TR 23735) \cite{webISO237352019}} and \enquote{Methods for Evaluating Other Road User Behavior in the Presence of Automated Vehicle External Communication (ISO/AWI TR 23720) \cite{webISO237202019}.} J3134 provides the most detailed guidance on AV lamps to date, recommending turquoise as their colour~\cite{webSAEJ3134201905}. The standard limits itself to AV marker lamps, however, stating that more studies are needed to guide the design of AV yield signal lamps. Our work could contribute to the development of those standards.

\subsection{AV Visual Signals}
Vehicle lights (e.g., brake and turning signals) have long been used to communicate vehicles' impending actions. In the context of automated driving, Lagstrom \textit{et al.} was among the first to investigate AV visual signals as communication to pedestrians. Namely, they designed the Autonomous Vehicles' Interaction with Pedestrians (AVIP) system that uses different signal patterns to represent vehicle intentions, which is reported to \enquote{increase pedestrians' willingness to cross and make the crossing experience more pleasurable \cite{Lagstrom2016}.} However, in a Wizard-of-Oz study conducted on the streets of Arlington, Virginia~\cite{webFordVT}, researchers used visual signals to communicate vehicle intention but found them to have little effects on pedestrian behaviours. More research is needed to make sense of the discrepancy.

\section{System Prototype}

We design and build a visual communication system using WS2815 individually programmable LED lights. These LEDs are evenly spaced 16.6 mm apart on a flexible circuit strip and each LED pixel can emit the full range of $256^3$ RGB colours with an intensity of 1.95--3.5 candela (cd). The lights are integrated with the Autonomoose self-driving research platform owned by the University of Waterloo Intelligent System Engineering (WISE) lab. The platform is based on a 2014 Lincoln MKZ hybrid Sedan (Figure~\ref{fig:moose}). It is outfitted with lidar, inertial, GPS, and camera sensors, and controlled by an on-board computer running an automated driving software stack on the Robotic Operating System (ROS). The vehicle is capable of operating in SAE Level 3 automation~\cite{webSAEJ3016201806} on urban roads. The LED lights
are controlled by an Arduino Uno microcontroller, which is integrated with the vehicle's ROS software stack.

\begin{figure}[bht!]
  \centering
  \begin{subfigure}[b]{0.49\linewidth}
    \centering
    \includegraphics[width=\linewidth]{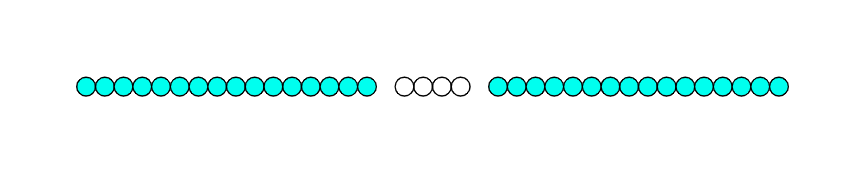}
    \caption{S-T: Solid Turquoise}
  \end{subfigure}
  \begin{subfigure}[b]{0.49\linewidth}
    \centering
    \includegraphics[width=\linewidth]{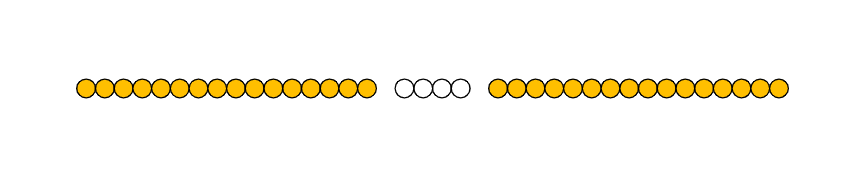}
    \caption{S-A: Solid Amber}
  \end{subfigure}  
  \begin{subfigure}[b]{0.49\linewidth}
    \centering
    \includegraphics[width=\linewidth]{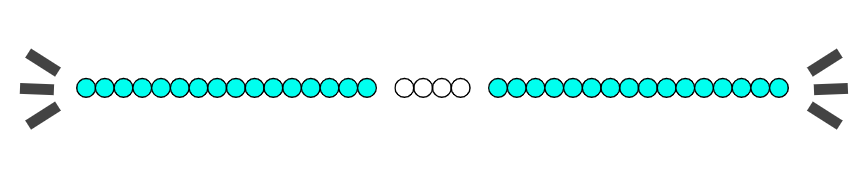}
    \caption{B-2: Blink (2 Hz)}
  \end{subfigure}
  \begin{subfigure}[b]{0.49\linewidth}
    \centering
    \includegraphics[width=\linewidth]{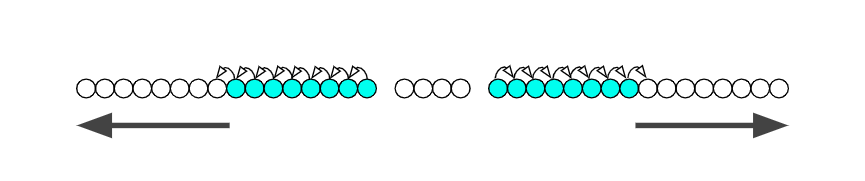}
    \caption{C-4: Chase (4 Hz)}
  \end{subfigure}  
  \begin{subfigure}[b]{0.49\linewidth}
    \centering
    \includegraphics[width=\linewidth]{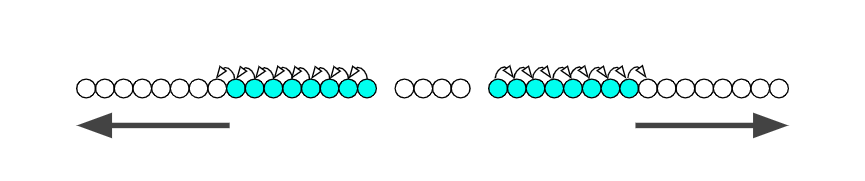}
    \caption{E-x: Expanding}
  \end{subfigure}
  \begin{subfigure}[b]{0.49\linewidth}
    \centering
    \includegraphics[width=\linewidth]{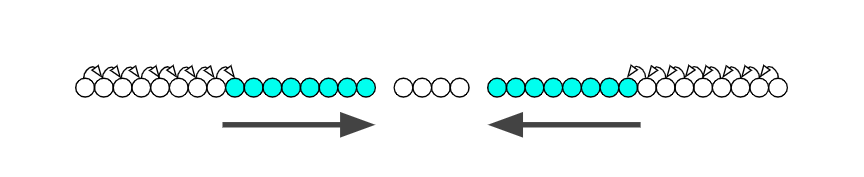}
    \caption{S-x: Shrinking}
  \end{subfigure}
  \caption{The six visual signals featured in this study}
  \label{fig:visual_signals}
\end{figure}

Using this setup, the research vehicle has direct control of the LEDs, which we programmed to display the following six representative visual signal patterns (Figure \ref{fig:visual_signals}): a) \textbf{S-T: Solid Turquoise}; b) \textbf{S-A: Solid Amber}; c) \textbf{B-2: Blink}; d) \textbf{C-4: Chase}; e) \textbf{E-x: Expanding}; f) \textbf{S-x: Shrinking}. S-T an S-A are static signals similar to a vehicle's stop light. B-2 is a dynamic signal that flashes continuously at 2 Hz. C-4 is also a dynamic signal that moves horizontally, expanding outward and resetting at 4 Hz. E-x and S-x correlate with a vehicle's speed and visualize its acceleration and deceleration, respectively. The studied pattern sample is necessarily limited, but includes stationary, flashing, and moving patterns. Our goal is to study how pedestrians interpret the vehicle's intentions based on these visual signal patterns and their features. 

Finally, the LED strips are mounted by group onto different parts of the vehicle. Group 1 (8 LED pixels, 13.28 cm in length) is designed as an ADS marker lamp; Group 2 through 5 (32 LED pixels, 53.12 cm in length) are designed as intention communication signals. 
We refer to Group 2 and 3 as \textbf{Top-Mount} visual signals because they appear directly above the vehicle's windshield (Figure~\ref{fig:moose}). By contrast, Group 4 and 5 are integrated into the vehicle's front grill; thus, we refer to them as \textbf{Front-Mount} visual signals. Group 1 collectively emits up to 28 cd, meeting the photometry requirements (2.5--300 cd in daytime and 0.5--125 cd in nighttime) for ADS Marker Lamp \cite{webSAEJ3134201905}.  

\section{Experiment 1}
In E1, we investigate how to design visual signals to enhance visibility in a pedestrian's peripheral vision. Specifically, we explore the situation where a pedestrian is crossing a street without looking for oncoming vehicles. It is a hazardous situation made worse by the increased usage of handheld electronic devices \cite{Hyman2009}. We hypothesize an AV can use visual signals to attract pedestrians' attention even when they are not looking directly at it, thereby mitigating this hazard. The experiment answers the questions: \textbf{RQ1}---What visual signal pattern is most noticeable in a pedestrian's peripheral vision? \textbf{RQ2}---Where should we place the visual signal to enhance its peripheral visibility?



\subsection{Methods and Procedure}
Thirty-one participants recruited from a university campus (22M/9F, ages 18-44) participate in E1. 51.6\% of them report crossing a street 6-10 times per day and 80.6\% hold a valid driver’s license. Each receives \$20 Canadian as a token of appreciation. E1 runs in a campus parking lot from 10:00 am to 10:30 pm, covering different times of the day (Figure~\ref{fig:E1_brightnessconditions}).

Participants complete an initial questionnaire (demographics), and are randomly assigned to go to one of four locations at an imaginary crosswalk perpendicular to the AV's path of travel. The vehicle, initially out of sight, approaches the crosswalk at 5km/h, while displaying one of the six randomly ordered visual signals; in the baseline trial, no visual signals are displayed. Participants are instructed to use peripheral vision, to keep their head straight and to say STOP when they spot the vehicle or its visual signal. At this point, a nearby researcher sends a Bluetooth signal to the vehicle to turn off the visual signal and record the distance gap between the vehicle and the participant. Seven trials (1 baseline and 6 experimental) are administered at each location and six visual signals are tested, i.e., S-T, B-2, C-4 top-mounted and front-mounted. We record the detection distance gap, the signal description, and the ambient illuminance (lux) level for each trial. Moreover, two practice trials are given prior to the actual experiment.


\subsection{Results}

\subsubsection{Signal Visibility}
Twenty-eight participants (N=28) complete E1, and the results reveal that dynamic signals (B-2 and C-4) are more visible in peripheral vision than a static signal (S-T). That is, the participants detect the dynamic signals earlier, as measured by the distance gap (in meters) between the vehicle and the participant when the participant detects the vehicle and says 'STOP'. 

This finding is consistent with previous optometry research suggesting that peripheral vision is good at detecting motion \cite{To2011}. Using the two factor Analysis of Variance (ANOVA) test, we compare the distance gaps produced by different visual signals and find the following: the distance gaps induced by the B-2 signal are significantly larger than the gaps induced by the S-T signal, (F(1,440) = 13.96, p $<$ 0.001, $\eta^2$ = 0.031); C-4 gaps are larger than S-T gaps, (F(1,440) = 14.8, p $<$ 0.001, $\eta^2$ = 0.032); and, there is no significant difference between the B-2 and C-4 gaps, (F(1,440) = 0.007, p = 0.934, $\eta^2$ $<$ 0.001). Specifically, the average distance gap induced by the B-2 signal is 7.95 meters, 7.99 meters for the C-4 signal, and 5.84 meters for the S-T signal.

The experiment also finds that mounting location, top- or front-mount, does not significantly impact the signals' visibility. Namely, the average distance gaps induced for the top- vs. front-mount S-T signals are 5.7 vs. 5.98 meters, (F(1,216) = 0.116, p = 0.734, $\eta^2$ $<$ 0.001); for top- vs. front-mount B-2 signals are 8.03 vs. 7.87 meters, (F(1,216) = 0.041, p = 0.841, $\eta^2$ $<$ 0.001; and for top- vs. front-mount C-4 signals are 7.78 vs. 8.21 meters, (F(1,216) = 0.322, p = 0.571, $\eta^2$ = 0.002).

\begin{figure}[tbh!]
  \centering
  \begin{subfigure}[t]{0.32\linewidth}
    \includegraphics[width=\linewidth]{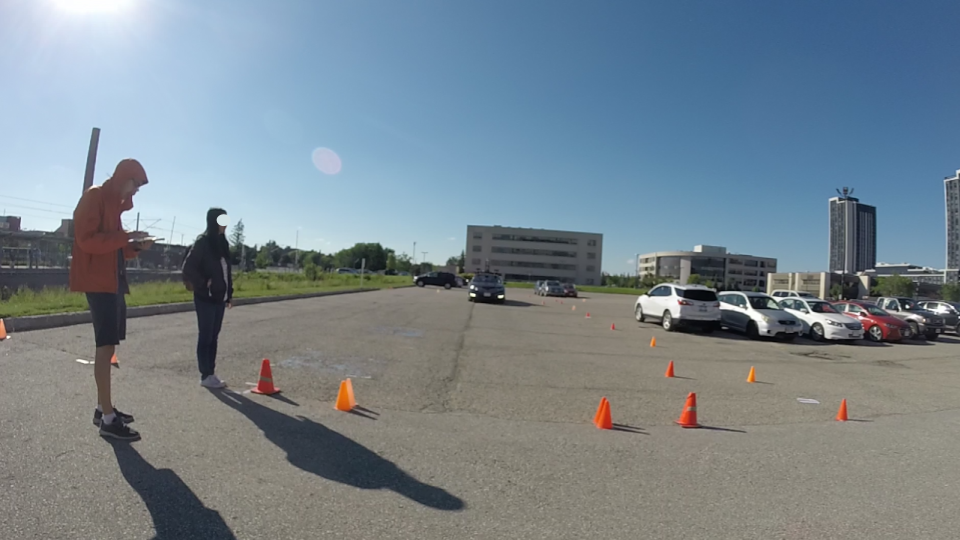}
     \caption{Sunny}
  \end{subfigure}
  \begin{subfigure}[t]{0.32\linewidth}
    \includegraphics[width=\linewidth]{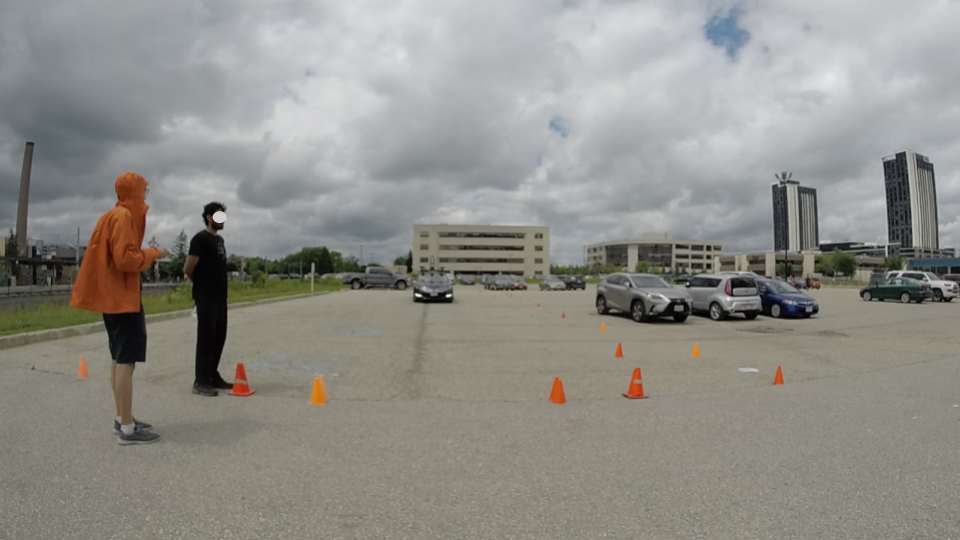}
    \caption{Overcast}
  \end{subfigure}
  \begin{subfigure}[t]{0.32\linewidth}
    \includegraphics[width=\linewidth]{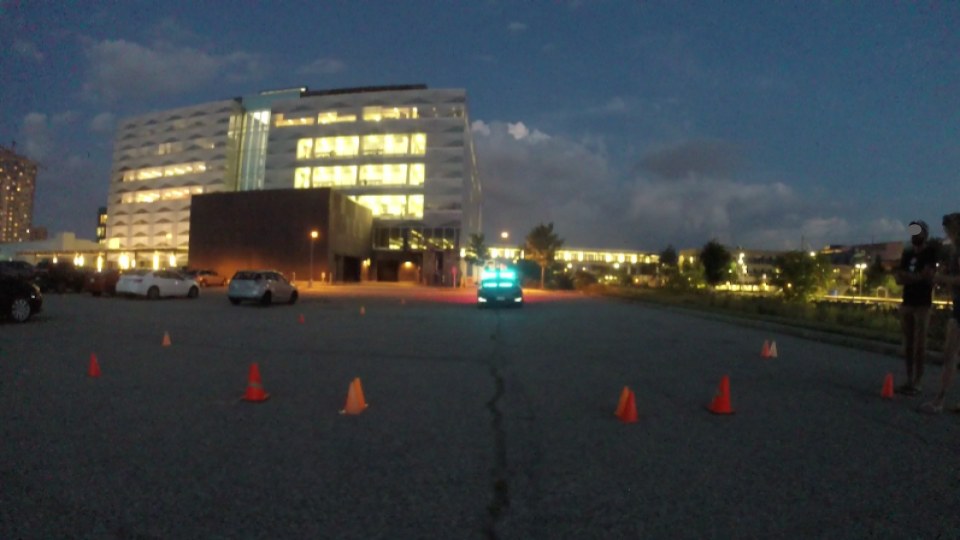}
    \caption{Nighttime}
  \end{subfigure}
  \caption{E1 snapshots of different ambient conditions}
  \label{fig:E1_brightnessconditions}
\end{figure}
\setlength{\textfloatsep}{0.3\baselineskip plus 0.2\baselineskip minus 0.5\baselineskip}
\subsubsection{Impacts of ambient illuminance}
\begin{figure}[tbh!]
  \centering
  \includegraphics[width=0.80\linewidth]{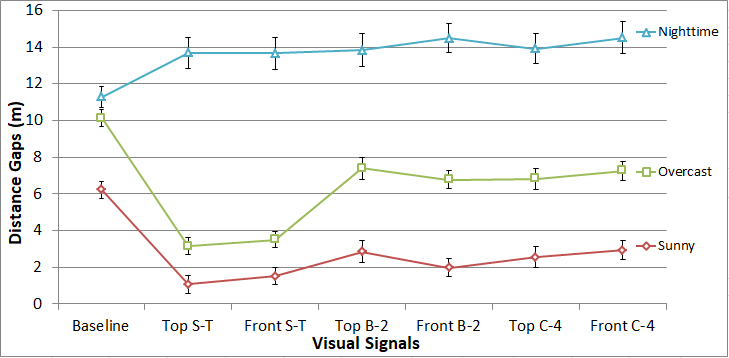}
  \caption{E1 average distance gaps (meters) induced by visual signals and ambient brightness and Standard Error of the Mean (S.E.M.) (N=8,6,11)}
  \label{fig:E1_brightnessgaps}
\end{figure}

Results confirm that the visibility of signal is heavily influenced by ambient brightness. We group the experiment results according to three ambient conditions: (1) sunny---lux level $>$ 10,000 (N=11); (2) overcast---lux between 1,000--10,000 (N=6); and (3) nighttime---lux $<$ 1,000 (N=8). Figure \ref{fig:E1_brightnessconditions} illustrates the difference in these conditions. Comparing the distance gaps, we find that signals are more visible during nighttime than in sunny conditions (see Figure \ref{fig:E1_brightnessgaps}). However, there are no significant differences among the distance gaps induced by the six experiment signals in nighttime, (F(5,168) = 0.187, p = 0.967, $\eta^2$ = 0.005), and sunny conditions, (F(5,120) = 1.62, p = 0.161, $\eta^2$ = 0.053). The visual advantage of the B-2 and C-4 signals are only significant in overcast conditions, (F(5,240) = 14.12, p $<$ 0.001, $\eta^2$ = 0.213). Finally, considering the baseline cases where visual signals are not used, participants are able to see the vehicle before the signals in sunny and overcast condition, while the reverse is true in nighttime condition.  

\subsection{Discussion}
E1 results demonstrate that a visual signal pattern involving {\it movement} is generally more visible than a static one. However, when we take ambient brightness into account, we see the visual advantage of moving signals only in overcast conditions. Namely, in sunny and nighttime conditions, the peripheral visibility of the three tested signal patterns are similar. Moreover, in sunny and overcast conditions, participants can see the moving vehicle before the visual signals, making visual signals less useful for a distracted pedestrian under these conditions. 
This is in contrast to nighttime, when participants can spot the signals before the vehicle. 

While visual signal mounting location does not significantly impact visibility, we argue that it is better to mount visual signals near the top of the windshield \cite{Lagstrom2016} because it would minimize the likelihood of signal obstruction by dirt and debris. Altogether, we derived the following two visual signal design recommendations from the experiment: \textbf{DR1}---Use moving or static visual signal patterns, such as Blink, Chase, and Solid, to enhance autonomous vehicle visibility during nighttime; \textbf{DR2}---Place vehicle signals near the top of the windshield to minimize visual interference.


\section{Experiment 2}
The goal of E2 is to learn the cognitive impact of visual signal patterns. Namely, we hypothesize that pedestrians share certain mental intuitions when it comes to interpreting visual signals, and the time a pedestrian needs to react to a signal is directly correlated to the mental effort necessary to process it. Hence, this experiment is designed to discover these intuitions and investigate whether training can override these intuitions, making it easier for pedestrians to make sense of the signals. Specifically, we investigate the following questions: \textbf{RQ3}---How would a pedestrian react to different visual signals intuitively? \textbf{RQ4}---How would a pedestrian react to different visual signals after training? \textbf{RQ5}---How long does it take for a pedestrian to interpret and react to different visual signals?


\subsection{Methods and Procedure}
E2 has same participants and location as E1. E2 begins with a \emph{baseline test} to measure participants' reaction time. Red and green signals resembling the traffic Stop and Go signal lights are shown on the vehicle. Then, participants are instructed to react to the signals by pressing a Stop or Go button on a custom Android app. The times for each decision are recorded. Two \emph{intuition tests} follow, with the vehicle stopped at the crosswalk in one test and 17 meters away in the other. In each test, the vehicle displays a randomized sequence of 18 signals, which consists of the six experiment signals (S-T, S-A, B-2, C-4, E-x, and S-x), each appearing 3 times. Without knowing what the signals mean, participants are asked to interpret them, as quickly as possible, and indicate their decision by pressing the Stop or Go button on the mobile app. 
Finally, we administer two {\it learned tests}, which follow the same procedure as the intuition tests except that participants are first given a two-minute training about the meaning of the visual signals. In this way, we can compare how training may impact pedestrians' behaviours. 


\subsection{Results}
\subsubsection{Intuitive and Learned Reactions}
\begin{figure}[tbh!]
  \centering
  \includegraphics[width=0.80\linewidth]{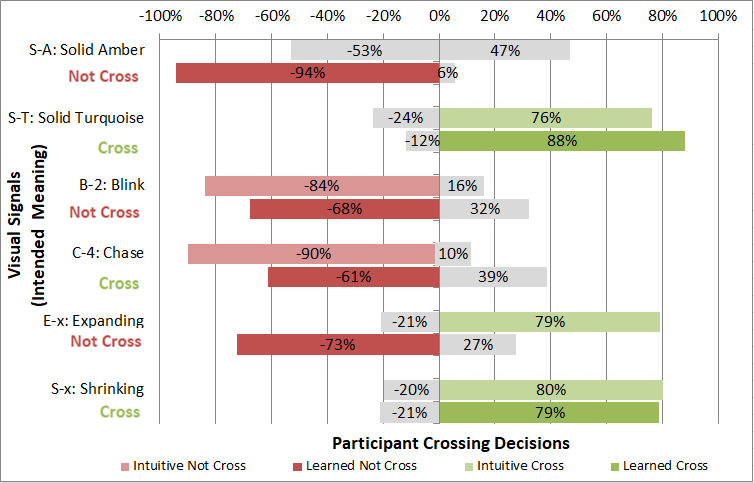}
  \caption{E2 participants' intuitive and learned reactions to visual signals (N=17)}
  \label{fig:E2_decisions}
\end{figure}

Seventeen participants (N=17) complete E2. Figure \ref{fig:E2_decisions} captures the experimental results---red columns mark the scenarios where a majority of participants decide not to cross; green indicates when the majority decide to cross; and grey indicates the decisions of the minorities. The Y-axis shows the names of the six visual signals and their intended meaning, as used in participant training. Overall, the results show that participants intuitively react to dynamic signals like B-2 and C-4 by not crossing, and they react to the S-T, E-x, and S-x signals by crossing. However, pedestrians are undecided about how to react to the static Amber S-A signal. By comparison, after training, pedestrians learn to react to the static Amber S-A and the S-x Shrinking signals by not crossing. However, training is unable to override participants' intuition to avoid crossing under C-4 signal.

\subsubsection{Decision time}
Table \ref{table:E2_time} captures participants' intuitive, learned, and average decision times to respond to the six experiment visual signals. Overall, pedestrians are quickest to respond to dynamic signals like B-2 and C-4 (M=2.5\,s). However, after training, they are quickest to respond to static signals like S-T and S-A (M=2.45\,s). The average decision time to the E-x and S-x signal are slowest (M=3.1\,s); the fact that these signals take longer to animate could be the reason. Moreover, the decision time for the S-A signal decreases from 2.7\,s to 2.3\,s after training, suggesting that training has removed the ambiguity and confusion about the signal. A similar observation can be made for the E-x signal, where participant decision time decreases from 3.2\,s to 2.8\,s. Coincidentally, the decision time for both signals is reduced by about 0.4\,s, which can be interpreted as a tangible benefit of training. However, training can also have an opposite effect on decision time. Namely, decision time for both the B-2 and C-4 increased by 0.4\,s and 0.5\,s, suggesting the need of mental effort to distinguish these two similar signals.

The vehicle's position---i.e., whether the vehicle is stopped in front of the crosswalk or 17 meters away---does not significantly change participants reactions and decision time, (F(1,1212) = 0.124, p = 0.725, $\eta^2$ $<$ 0.001). However, the ANOVA test confirms that there exists significant interaction between training and decision time, (F(5,1212) = 3.8, p $<$ 0.01, $\eta^2$ = 0.015); and, decision times are significantly different for each visual signal, (F(5,1212) = 5.3, p $<$ 0.001, $\eta^2$ = 0.021).

\begin{table}[bth!]
\centering
\captionsetup{justification=centering}
\begin{tabular}{||P{0.8cm} P{0.8cm} P{0.8cm} P{0.8cm} P{0.8cm} P{0.8cm} P{0.8cm}||} 
 \hline
  & S-T & S-A & B-2 & C-4 & E-x & S-x \\ [1ex] 
 \hline\hline
  Average:  & 2.8 & 2.5 & 2.9 & 2.6 & 3.0 & 3.1 \\
  Intuitive:  & 2.9 & 2.7 & 2.7 & 2.3 & 3.2 & 3.2 \\
  Learned: & 2.6 & 2.3 & 3.1 & 2.8 & 2.8 & 3.0 \\
  \hline
  Difference: & -0.3 & -0.4 & +0.4 & +0.5 & -0.4 & -0.2 \\
 \hline
\end{tabular}
\caption{E2 participants' intuitive and learned decision time to visual signals (seconds) (N=17)}
\label{table:E2_time}
\end{table}

\subsection{Discussion}
E2 revealed that visual signal features can cause pedestrians to react to signals in different ways. Namely, the frequency of a signal is one of those influencing factors. For example, participants suggest that the \enquote{fast} frequency of the B-2 and C-4 signals convey a sense of urgency and caution, similar to a flashing pedestrian crossing light; in this case, pedestrians tend to react to this kind of signals by stopping to cross. On the other hand, the \enquote{slow} frequency of the E-x and S-x conveys a sense of calm, making it more likely for pedestrians to decide to cross. Participants mention colour as the main determining factor for how they perceive the solid signal patterns. For example, they perceive the novel Turquoise colour signal to be calm and mellow, implying that it is OK to cross. However, they are confused about the Amber colour signal because they do not know whether the signal is indicating the vehicle's state of operation or informing pedestrians which action to take. Finally, results show that participants tend to be risk-averse. Training is not able to overcome participants' intuition about the C-4 signal, which they perceive as danger. Pedestrians are not willing to risk their lives against their intuition.

Regarding decision time, the experiment shows that participants react intuitively and quickly to Blink and Chase signals, perhaps due to our given instinct to avoid danger. However after training, the two Solid signals become easiest to interpret and decision time decreases by 453\,ms, which confirms that our minds are quicker to process colour than movement \cite{Moutoussis1997}. On the other hand, decision time to Blink and Chase signals increases by 433\,ms after training, reflecting the confusion that participants have about the two signals. Finally, participants generally take longer to react to the slow moving Expanding and Shrinking signals, namely, 440\,ms longer compared to the Solid signals, suggesting the animation speed of visual speeds should be considered in their design. In combination, we derive the following design recommendations: \textbf{DR3}---Use fast-moving visual signal patterns to communicate urgency and danger to deter pedestrians from crossing; \textbf{DR4}---Use slow-moving visual signal patterns to communicate calm and safety to encourage pedestrians to cross; \textbf{DR5}---Do not use slow-moving visual signal patterns to deter pedestrians from crossing; it may endanger them; \textbf{DR6}---Validate any visual signal pattern with pedestrian's intuition; do not work against it; \textbf{DR7}---Use solid visual signal patterns to reduce pedestrians cognitive load and to make their crossing decision easier. 


\section{Experiment 3}
E3 is a novel AV-pedestrian communication experiment using an AV in automated mode. It explores whether participants will cross the street when an AV is approaching or is stopped at the crosswalk. We evaluate the effectiveness of visual signals by asking the following questions: \textbf{RQ6/7}---What visual signals would encourage or deter a pedestrian from crossing in front of an approaching/a stationary AV? 
\textbf{RQ8}---How does repeated exposure to visual signals influence a pedestrian's crossing decision? \textbf{RQ9}---How does familiarity and trust of an AV influence a pedestrian's crossing decision?


\subsection{Apparatus}
Existing AV-pedestrian interaction experiments fall into three categories: (1) Conceptual approach in which participants decide what they will do in hypothetical scenarios \cite{Mahadevan2018}, (2) Simulation approach in which participants interact with vehicle in a virtual immersive environment \cite{Jayaraman2018}, and (3) Wizard-of-Oz approach in which participants interact with a supposedly autonomous vehicle driven by a concealed driver \cite{Rothenbucher2016}. 
E3 takes a new approach in which participants interact with a real AV. To this end, we program the Autonomoose vehicle to execute two automated scenarios. (1) \textbf{Drive-By Scenario}: the AV approaches a crosswalk at 40 km/h, and at 100 meters displays a visual signal indicating it will not stop. Keeping the same speed and signal, it then passes the crosswalk. (2) \textbf{Stop-and-Go Scenarios}: in scenario 2a---the AV approaches a crosswalk at 40 km/h, and at 100 meters displays a visual signal indicating it will stop. Then, after 5s when it is 48 meters from the crosswalk, it starts to decelerate gradually and stops in front of the crosswalk. In scenario 2b---the AV displays the S-T signal to indicate it is at rest; after 10s, it displays a visual signal to indicate its intention to accelerate. After another 10s, the vehicle accelerates and drives pass the crosswalk. 
While the AV executes these scenarios autonomously, a safety driver is behind the wheel and ready to take control at all times. Moreover, the participants were prohibited from coming within 3 meters of the vehicle's lane of travel. Instead, participants were taught to use two arm signals to indicate their crossing decision: 1) they extend their right arm to indicate a willingness to cross and 2) they bend their arm to point at the sky to indicate they are not willing to cross. 
A researcher on board the vehicle watches the participants and records their crossing decision by pressing an electronic button, which is connected to 
the vehicle's software stack. The stack then logs the participant's decision along with vehicle's position, current maneuver, and the visual signal state. We extract the experiment results from these log files.

\begin{figure}[bth!]
  \centering
  \begin{subfigure}[t]{0.32\linewidth}
    \includegraphics[width=\linewidth]{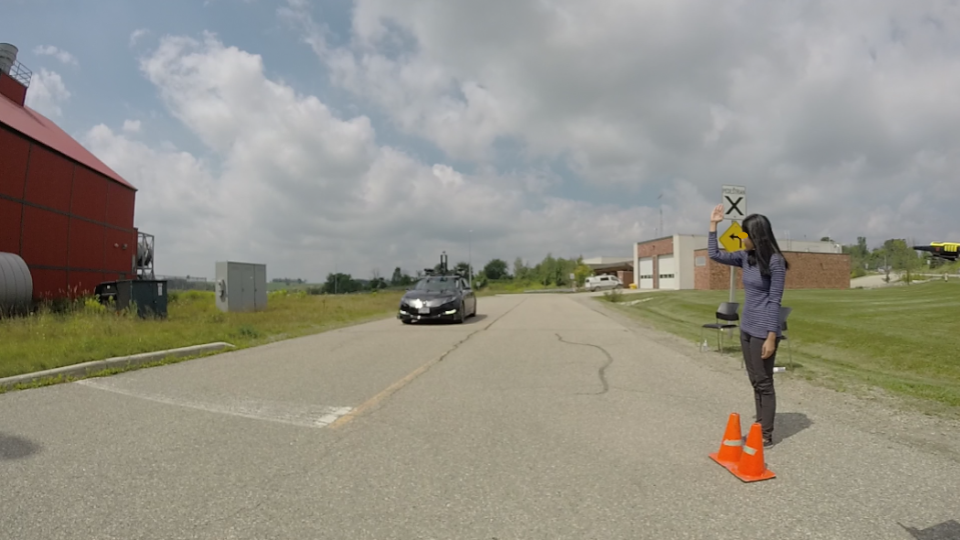}
    \caption{Drive-By scenario --- AV driving pass}
  \end{subfigure}
  \begin{subfigure}[t]{0.32\linewidth}
    \includegraphics[width=\linewidth]{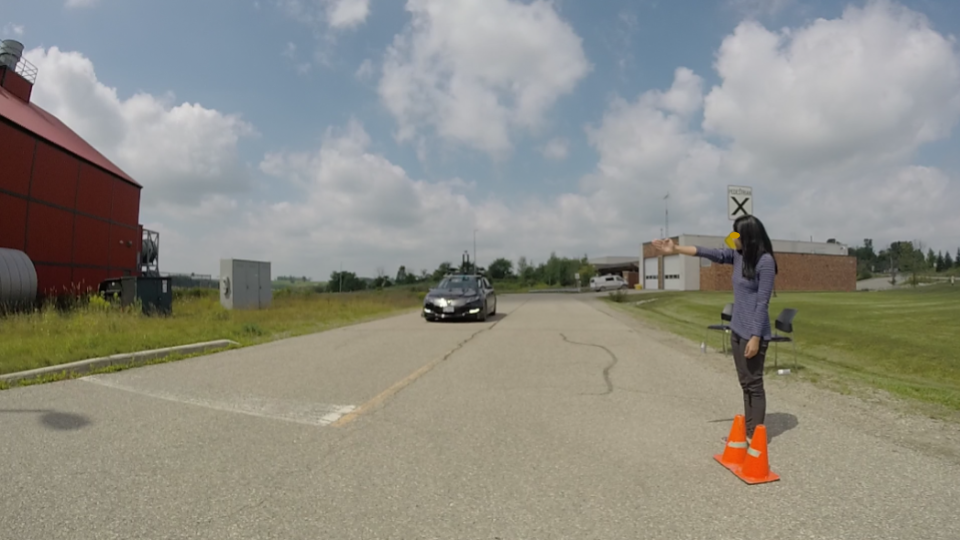}
     \caption{Stop-and-Go scenario --- AV approach}
  \end{subfigure}
  \begin{subfigure}[t]{0.32\linewidth}
    \includegraphics[width=\linewidth]{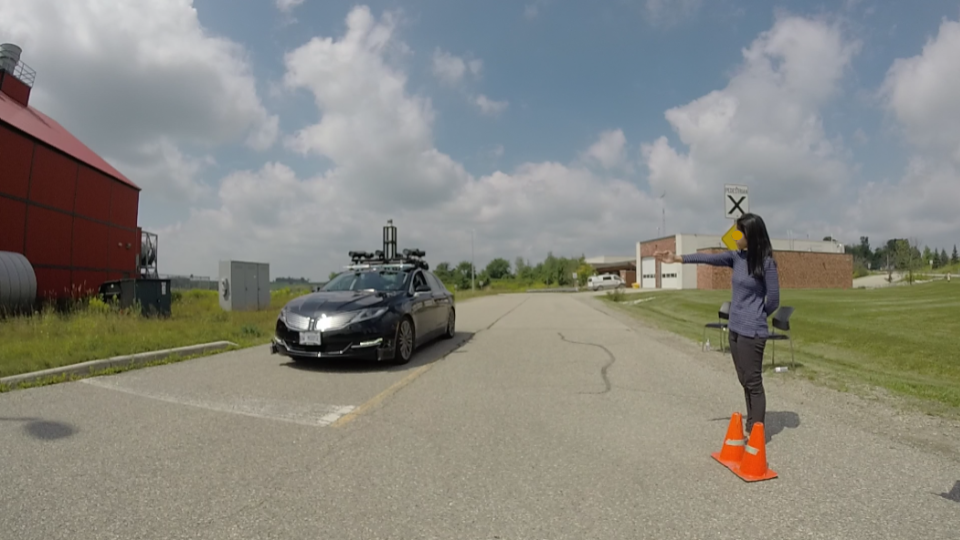}
    \caption{Stop-and-Go scenario --- AV stopped}
  \end{subfigure}
  \caption{E3 snapshots of automated scenarios}
  \label{fig:E3_situations}
\end{figure}
\vspace*{-\baselineskip} 


\subsection{Methods and Procedure}
E3 has 22 participants recruited from a university campus (14M/8F, ages 18-44). 40.9\% of them report crossing a street 6-10 times per day and 63.6\% hold a valid driver’s license. They receive \$20 Canadian as appreciation. The experiment runs at the Waterloo Region Emergency Services Training and Research Centre (WRESTRC) test track, 9:00 am to 5:00 pm.
Similar to E1, participants complete a demographic questionnaire. Then, about half of them receive an autonomous ride around the test track, to demonstrate the vehicle's self-driving capability.  The experiment starts with a participant standing at the entrance of an imaginary crosswalk, followed by the AV approaching autonomously and executing either the Drive-by or the Stop-and-Go scenario. The vehicle communicates its intention by using a visual signal; in response, the participant is asked to indicate their decision to cross or stop crossing at each instant.

In general, each participant completes two repeated runs of six trials, which consists of 1 baseline Drive-By trial without visual signal, 2 Drive-By trials with the S-A and B-2 signals, 1 baseline Stop-and-Go trial without visual signal, and 2 Stop-and-Go trials showing the S-T and S-x signals when the AV is approaching and the S-A and B-2 signals when the AV is about to accelerate from rest. The order of the trials and visual signals are randomized. 


\subsection{Results}

\subsubsection{Effect of Visual Signals}
Nineteen participants (N=19) complete scenario 1 and 2a. The results in Table~\ref{table:E3_approach} show that visual signals (S-T, S-x, S-A, B-2) do not significantly encourage or deter a pedestrian from crossing in front of an approaching vehicle whether it is stopping or not, (F(5,195) = 0.26, p = 0.933, $\eta^2$ = 0.007). In contrast, fourteen participants (N=14) complete scenario 2b and the results show that the signals (S-A, B-2) can deter pedestrians from crossing in front of a stationary vehicle. Namely, in the absence of a visual signal (baseline case), most participants continue to cross in front a stationary vehicle until it starts to move. The average stop crossing time for the baseline case is 18.84\,s (measured from the start of 2b); interestingly, this time is short of the 20\,s mark when the vehicle starts to move, because four participants decide to stop crossing just before and in anticipation of the movement. However, when visual signals (S-A and B-2) are used to communicate the AV's intention to accelerate, participants respond to the visual signals and stop crossing much earlier, at 14.79\,s and 15.76\,s, respectively. These are statistically significant results compared to the baseline trials, (F(1,49) = 19.62, p $<$ 0.001, $\eta^2$ = 0.286) and (F(1,52) = 12.31, p $<$ 0.001, $\eta^2$ = 0.191). Thus, visual signals can prompt pedestrians to yield to a waiting AV at a crosswalk. 
 
\begin{table}[htb!]
\centering
\captionsetup{justification=centering}
\begin{tabular}{||P{0.8cm} P{0.8cm} P{0.8cm} P{0.8cm} P{0.8cm} P{0.8cm} P{0.8cm}||} 
 \hline
 \multirow{2}{*}{} &
 \multicolumn{3}{c}{\underline{Stop-and-Go}} &
 \multicolumn{3}{c||}{\underline{Drive-By}} \\
  & {Baseline} & {S-T} & {S-x} & {Baseline} & {S-A} & {B-2} \\ [1ex] 
 \hline\hline
  {Average}: & 33.72 & 31.15 & 33.98 & 30.87 & 31.36 & 34.44 \\
 \hline
\end{tabular}
\caption{E3 average distance gaps (meters) induced by visual signals as an AV is approaching (N=19)}
\label{table:E3_approach}
\end{table}


\subsubsection{Impacts of Learning}
Seventeen participants (N=17) completed two experiment runs. However, comparison of distance gaps and stop crossing time between the two runs shows no significant difference in the participants' crossing behaviours, (F(1,174) = 0.011, p = 0.915, $\eta^2$ $<$ 0.001) and (F(1,68) = 0.679, p = 0.413, $\eta^2$ $<$ 0.01), suggesting that no meaningful learning has occurred. Future work should study whether more exposure can improve learning.

\subsubsection{Impacts of Trust}
\begin{figure}[tbh!]
  \centering
  \includegraphics[width=0.9\linewidth]{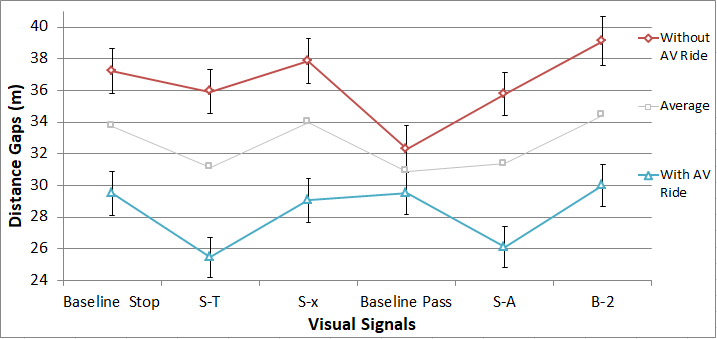}
  \caption{E3 average distance gaps (meters) induced by visual signals and trust of the vehicle and S.E.M. (N=10,9)}
  \label{fig:E3_trustgaps}
\end{figure}

In total, nine participants (N=9) receive a ride in the AV prior to experiment and ten do not (N=10). Comparison of distance gaps between the two groups have significant differences in their crossing behaviour, (F(1,119) = 10.69, p $<$ 0.01, $\eta^2$ = 0.051). Namely, participants with the autonomous ride experience become more risk-tolerant, regardless of whether visual signals are used.  This finding may be attributed to the fact that participants become more confident about the AV's capability after having witnessed it first-hand, and such trust can influence pedestrians' crossing behaviours. This finding aligns with the study by Rossi \textit{et al.}, which suggests people's trust for a robot increases with their awareness of the robot's design capabilities \cite{Rossi2018}. Figure \ref{fig:E3_trustgaps} illustrates the findings.


\vspace*{-\baselineskip} 

\subsection{Discussion}
E3 reveals that visual signals can convince a pedestrian to yield to a stationary AV at a crosswalk. Specifically, we find that the static S-A signal is just as effective as the dynamic B-2 signal at influencing pedestrians behaviour; post-experiment interviews reveal two potential explanations of this phenomenon: (1)  some participants suggest it is the sudden change of signal, i.e., from S-T to S-A or B-2, that communicated \enquote{the vehicle is about to do something different.} (2) some participants say that they react to the signals simply because they can see them better in close proximity. On the other hand, visual signals in scenario 2a are not enough to overwrite vehicle speed and distance as the dominating factors influencing a pedestrian's crossing decision \cite{Risto2017}. Another explanation could be that pedestrians fail to see the visual signals clearly at a distance. 

Finally, the findings suggest that pedestrians will learn to trust AVs with increasing familiarity. This finding is encouraging as the proliferation of AVs continues. However, it is also concerning, as previous human-robot interaction studies show that people can over-trust a robot and act irrationally as a result \cite{Salem2015}. Likewise, we observed such irrational behaviours in E3. Namely, the S-A signal is supposed to deter participants from crossing in front of an approaching AV. While participants who do not receive an autonomous ride understand and comply with the signal, participants who \enquote{trust} the vehicle do not, i.e., they erroneously assume that both the S-A and S-T signals mean that the vehicle has seen them and will stop. Assumptions like this and overconfidence in AVs could have catastrophic consequence for pedestrian safety in real traffic situations. Therefore, we recommend not to use static signal patterns like S-A to deter pedestrians to cross because they are prone to mental biases and misjudgements. Instead, we suggest using a moving visual signal pattern like S-x to encourage more careful deliberation before crossing. 
In summary, we derive the following design recommendations: \textbf{DR8}---Incorporate an abrupt change of visual signal patterns on a stationary AV to suggest to pedestrians that the vehicle is about to accelerate; \textbf{DR9}---Avoid using static visual signal patterns to deter pedestrians to cross.

\section{Experiment 4} 
The goal of E4 is to observe how AV visual signals work in real-traffic situations. In particular, we drive a Wizard-of-Oz AV on a public road and compare how pedestrians interact with it with and without visual signals. In doing so, we answer the following questions: \textbf{RQ10}---How does a pedestrian identify and recognize an AV? \textbf{RQ11}---Can pedestrians notice a visual signal on an AV? If so, what do they think about it? \textbf{RQ12}---Will pedestrians react differently to an AV equipped with visual signals? 

\subsection{Apparatus}
In preparation, we tint the windows and windshield of the Autonomoose vehicle with a Gila window tint that allows only 2.5\% Visible Light Transfer. In doing so, we conceal the driver and meet the university's safety standards. 
We use the following visual signals for E4: (1) ADS marker---to indicate the AV is operating in autonomous mode, (2) S-x shrinking---to indicate the AV is decelerating, (3) S-T Solid Turquoise---to indicate the AV is at rest, and (4) B-2 Blink---to notify pedestrians the AV's intention to accelerate.


\subsection{Methods and Procedure}
E4 has two types of participants. First, involuntary ones are those who happened to crossed paths with the AV. Their crossing behaviours are recorded and analyzed by the researchers. Secondly, voluntary ones are those who crossed path with AV and are willing to complete a questionnaire about their interaction. E4 has fifteen voluntary participants (14M/1F, ages 17-24). 40\% of them report crossing a street 11-15 times per day and 93.3\% hold a valid driver’s license. They are offered \$5 as a token of appreciation.

The experiment runs at a crosswalk on Ring Road (the main road at University of Waterloo), 9:00--11:30 am. 
Disguised as a self-driving vehicle, we drive passed the crosswalk a total of 24 times from each direction. Before the experiment, we have identified braking locations along the route, 30 meters before the crosswalk, at which place the driver would activate the S-x signal to communicate the AV is stopping. The deceleration time is choreographed to be 5s. Once the AV is stopped, the driver waits for pedestrians to cross, and after 5s, he activates B-2 signal to indicate the vehicle's intent to accelerate. Finally, when the crosswalk is clear, the AV accelerates and passes the crosswalk. 
The first 12 passes of the crosswalk are baseline trials, in which visual signals are not used; the subsequent 12 passes are experimental trials, in which signals are used to communicate intent. An on-board and an external camera capture the trials from first-person and third-person perspectives. In consideration of privacy, signs are posted along the Ring Road and on the vehicle to inform pedestrians about the study and about the possibility of being filmed. Pedestrians are given an option to remove themselves from the recording. The video recordings are reviewed by a researcher to analyze pedestrian behaviours and interaction with the vehicle.

\subsection{Results}
\subsubsection{AV and Signal Recognition}
In the questionnaire responses, voluntary participants consider visual signals and cameras to be the most recognizable features on the AV. Namely, after crossing path with the AV, 5 of 7 (71.4\%) pedestrians recall cameras as the most noticeable feature on the AV in baseline trials; In contrast, when visual signals are used, 4 of 8 (50.0\%) pedestrians recall cameras and 4 of 8 (50.0\%) recalled lights. Among them, two of the pedestrians recall both. 

As for interpretation of the signals, all four pedestrians who recall them assume the signals are meant \enquote{to indicate the intentions of the vehicle.} Nonetheless, they do not understand how to interpret or react to them. This could be the reason why the signals do not significantly improve pedestrians' crossing experience. For example, 4 of 7 (57.2\%) pedestrians from the baseline group and 4 of 8 (50.0\%) pedestrians from the signal group felt comfortable and safe crossing in front of the vehicle (6 or higher on a 7-point Likert scale). 

\subsubsection{Pedestrian behaviour with AV}
Analyzing video recordings, we observe most pedestrians exhibiting inattentiveness and carelessness while crossing in front of the AV. Out of the 52 baseline group pedestrians who cross paths with the AV, only 14 (26.9\%) look directly at the vehicle. Similarly, out of the 114 experimental group pedestrians who cross paths with the AV, only 31 (27.2\%) look directly at it. Moreover, 7 of 52 baseline group pedestrians (13.5\%) and 6 of 114 (5.3\%) experimental group pedestrians are distracted by various activities---e.g., phone use (most common), reading, or taking off jacket---while crossing. 

Comparing the reactions of those who looked, 7 of 14 (50.0\%) baseline group pedestrians change their crossing behaviours after looking at the AV. Namely, 4 (28.6\%) hesitate before crossing, 2 (14.3\%) yield to the vehicle, and 1 (7.1\%) starts jogging across the crosswalk. The rest of the pedestrians do not change their crossing behaviours. Similarly, 14 of 31 (45.2\%) experimental group pedestrians change their crossing behaviours after looking at the AV with visual signals. Namely, 3 (9.7\%) hesitate before crossing, 3 (9.7\%) yield to the AV, 4 (12.9\%) start jogging across the crosswalk, and 4 (12.9\%) stop after crossing and take photos of the AV. Overall, the two groups of pedestrians react similarly to the AV, except in two instances in experimental trials when the AV is displaying the B-2 signal, the four pedestrians who take photos of the vehicle exhibit great enthusiasm, which is not observed during baseline trials. However, it is unclear whether their reactions are induced by the visual signals.

\subsection{Discussion}

Overall, E4 results show that not all pedestrians can recognize whether a vehicle is autonomous and hence E4 affirms the SAE recommendation for AVs to use a marker light to indicate it is in ADS mode \cite{webSAEJ3134201905}. However, our study also reveals that a lack of awareness and recognition is one of the challenges of using AV visual signals in the wild. For example, the questionnaire study shows only 12.5\% of pedestrian think the visual signals are noticeable and none of them know what the signals mean; these could be the reasons why visual signals have been largely ineffective in public experiments such as the Ford study \cite{webFordVT}. These results also suggest that AV visual signals standardization and public education is required before the signals can be effective. 
In addition, the observation study reveals that herd behaviours are prevalent among pedestrians. Namely, in a group crossing situation, most pedestrians do not brother to look at the traffic if there are one or more pedestrians already in the crosswalk. Pedestrians seem to share an unspoken code of conduct for crossing---e.g., the first and last person to cross has the responsibility to look out for traffic; however, if someone else is already on the crosswalk, it is safe to cross because their presence implies safety. 
These norms seem to describe the majority of the pedestrian crossing behaviours observed in the experiment. While the majority of pedestrian-AV interaction studies focus on single pedestrians, our experiment demonstrates the need to conduct AV interaction studies with multiple pedestrians. 

We derived the following design recommendations from E4: \textbf{DR10}---Develop and use standardized visual signals given the potential of pedestrians being confused by the signals. \textbf{DR11}---Use marker lamps, such as those recommended by the SAE, to differentiate an AV from a manually-operated one. \textbf{DR12}---Educate the public about the existence of visual signals and their meanings.


\section{Conclusion}
The experimental results show that the role and effectiveness of AV visual signals depend on the situation, signal design, and pedestrians' familiarity with the signals and AV technology.  
Visual signals can convince a pedestrian to yield to a stationary AV at a crosswalk. In contrast, they have little impact on pedestrian behavior in front of an approaching AV: the herd behavior to follow other pedestrians already crossing, without looking at the vehicle, is prevalent, and for those who look, the vehicle’s deceleration profile is the dominating factor influencing their behavior. Related to signal design, we have also learned about the importance of not working against existing intuitions. These intuitions include high-frequency dynamic patterns indicating urgency and inducing caution and a sudden signal appearance change indicating intention change. Conversely, slow-moving signals turn out to convey a sense of calm, and thus should not be used to deter from crossing. We also see the opportunity to use color to reduce cognitive load, and the need to consider the strong impact of ambient illumination on signal visibility. Surprisingly, E3 reveals a significant impact of familiarity and trust of AVs on crossing behaviors. Consequently, efforts designed to educate the public about AVs to reduce the risk of overconfidence are essential. Finally, the field experiment results also underline the importance of standardizing AV signals and educating the public about them. For more details on the experiments, see~\cite{mst}.

\end{document}